\documentclass[runningheads]{llncs}
\usepackage[T1]{fontenc}
\usepackage{xurl}
\usepackage{hyperref}
\usepackage{tabularx}
\usepackage{amsmath}
\usepackage{amssymb}

\usepackage{array}
\usepackage{graphicx}
\graphicspath{
	{figures_main/}
}

\begin{document}
\title{A Time-Synchronous Prequential Evaluation Framework for Next-Week Student Disengagement Prediction from Moodle LMS Logs \vspace{-4mm}}
\titlerunning{Prequential Evaluation for Next-Week Student Disengagement}
%
%
\author{
David Otoo-Arthur\inst{1}\orcidID{0000-0003-2057-088X} \and
Terence L. van Zyl\inst{2}\orcidID{0000-0003-4281-630X}
}
\authorrunning{D. Otoo-Arthur and T. L. van Zyl}

\institute{
University of the Witwatersrand, Johannesburg, South Africa\\
\email{david.otoo-arthur1@students.wits.ac.za}
\and
University of Johannesburg, Johannesburg, South Africa\\
\email{tvanzyl@uj.ac.za}
}
\maketitle              

\begin{abstract}

Retrospective evaluation of early-warning models may not reflect information available when predictions are issued and outcomes become observable. We developed a time-synchronous prequential framework for online prediction of next-week student disengagement, operationalised as inactivity onset, from Moodle logs. Ordered by calendar time, models were updated only with available outcomes and remained unchanged for all predictions at each prediction cutoff; development, calibration and evaluation were chronologically separate. We compared stochastic gradient descent logistic regression (SGD-LR), online sequential extreme learning machine (OSELM) and echo state network (ESN) across predictive performance, fixed, frozen and rolling threshold policies, calendar-time variation and four behavioural feature sets. Held-out evaluation comprised 49,329 predictions with 37.68\% inactivity-onset prevalence. Discrimination was similar (average precision 0.665--0.679; ROC-AUC 0.755--0.774). OSELM had the highest recall, SGD-LR the highest precision and specificity, and ESN the highest Matthews correlation coefficient and balanced accuracy. Lower thresholds increased recall and alert volume, whereas rolling adaptation did not improve MCC over frozen thresholds. Weekly performance and calibrated-probability distributions varied without sustained deterioration. During model selection, cumulative-and-temporal features achieved the highest individual-set AP, with modest gains from combining all feature groups. These findings indicate that streaming early-warning evaluation should consider calibration, threshold-dependent performance, alert burden and calendar-time variation alongside aggregate discrimination.

	
\end{abstract}

\section{Introduction}
\label{sec1}

Student disengagement remains a concern in higher education because declining participation can precede poor academic outcomes and withdrawal~\cite{MorenoMarcos2020Persistence,Xia2023dropout}. Learning management systems (LMSs) record behavioural traces such as clickstream activity, resource access and assessment participation. Siemens~\cite{Siemens2013} and Baker and Inventado~\cite{Baker2014} identify these traces as an important source of evidence for learning analytics, and subsequent studies have used them to identify learners at risk of disengagement~\cite{Baker2020clickstream,Waddington2016EWS,RomeroVentura2020}.

Most dropout and disengagement studies, however, use static train--test partitions, cross-validation or retrospectively constructed course records~\cite{Prenkaj2020SurveyDropout,Gardner2019MOOC,Alghamdi2025MOOCReview,Marcolino2025MoodleDropout}. Such designs do not reproduce an operational setting in which observations arrive sequentially, next-week outcomes become available after the prediction horizon and models are updated only from observable outcomes. Repeated learner records and cumulative predictors also allow future information to enter feature construction, preprocessing, model selection, threshold calibration or evaluation~\cite{Kaufman2012,Varma2006Bias,Roberts2017CrossValidation}. Forward-only features and learner-aware partitions can address some forms of leakage~\cite{Paz2025CAPIRE}, but neither alone aligns prediction, delayed outcome availability and model updating. Temporal changes in participation may also reduce the reliability of performance estimates obtained from purely offline evaluation~\cite{Fei2015temporal,Liu2020temporal,Xia2023dropout,Labba2024towards}.

Prequential evaluation assesses models in the order predictions are issued and outcomes become observable. Dawid~\cite{Dawid1984Prequential} established this predict--observe--evaluate principle, while Gama et al.~\cite{Gama2013StreamEvaluation} developed procedures for evaluating learning algorithms on evolving data streams. Educational applications have, however, mainly addressed final academic outcomes using learner-level streams or coarse monthly, trimester and semester periods~\cite{ZavaletaSanchez2022Dynamic,Casalino2023Incremental,Casalino2024Time,Casalino2025Evolving}. Studies predicting dropout one week ahead have used preceding or cumulative activity, but have generally relied on retrospective partitions, fixed models, or independently trained week-specific models~\cite{Kloft2014WeeklyDropout,Xing2016TemporalDropout,XingDu2019DeepDropout,Alamri2021NextWeek}.

Consequently, a methodological gap remains in the time-synchronous evaluation of continuously updated models for next-week prediction across asynchronous educational data streams. Learner--course observations indexed by course-relative week may occur at different calendar times, while each outcome becomes available only after its one-week prediction horizon ends. Without enforcing these timing relationships, a model may be updated using outcomes unavailable when the prediction was made. Updating between observations predicted at the same time would also allow later predictions to use information unavailable to earlier ones, making the results depend on processing order. These conditions can compromise temporal validity and produce performance estimates that do not represent prospective deployment.

To address this problem, this study presents a time-synchronous prequential evaluation framework for next-week student disengagement prediction across asynchronous learner--course streams. Within this framework, learner--course observations are ordered by prediction time, and the model is updated only with outcomes already known. It then remains fixed for all predictions made at the same time. The data were divided chronologically so that all model-development decisions were made before the final evaluation period. This design evaluates adaptive models under the information and timing constraints of a prospective early-warning system. The evaluation compares three sequentially updated models: stochastic gradient descent logistic regression (SGD-LR), online sequential extreme learning machine (OSELM) and echo state network (ESN), representing linear, nonlinear feed-forward and recurrent approaches.

\subsection{Contributions}

Our study makes three contributions. First, we introduce a time-synchronous prequential evaluation framework for asynchronous learner--course streams in which the model is updated only with available outcomes and remains unchanged for predictions made at the same time. Second, we compare three streaming models and the predictive value of four feature sets under the same chronological design. Third, we examine how threshold policies affect which learners are flagged and how many alerts are produced, and whether predictive performance and calibrated probabilities vary over calendar time.

\section{Related Work}
\label{sec:related_work}

\subsection{Next-Week Disengagement Prediction}

Next-week dropout prediction aims to identify learners early enough for intervention~\cite{Kloft2014WeeklyDropout,Xing2016TemporalDropout,XingDu2019DeepDropout,Alamri2021NextWeek}. Early studies showed that activity from preceding course weeks could predict dropout in the following week. Kloft et al.~\cite{Kloft2014WeeklyDropout} used clickstream features with linear support vector machines, training separate models for each course week and evaluating them through five-fold cross-validation. Subsequent studies represented activity over time: Xing et al.~\cite{Xing2016TemporalDropout} combined information from preceding weeks through temporal stacking, while Xing and Du~\cite{XingDu2019DeepDropout} examined cumulative weekly features and compared deep learning with conventional classifiers.

Later work extended next-week prediction across multiple courses and course runs. Alamri et al.~\cite{Alamri2021NextWeek} incorporated timing, jumping and catch-up behaviours across seven courses and 29 course runs, training on earlier runs and evaluating on a later run. This provided stronger separation between model development and evaluation. Together, these studies established the relevance of a one-week horizon and the value of representing prior activity over time. However, they generally relied on retrospective datasets and either trained week-specific models or kept models fixed during evaluation. They therefore did not examine continuously updated models when outcomes become available later and records must be ordered by calendar time.

\subsection{Prequential Evaluation in Educational Prediction}

Prequential evaluation follows a predict--observe--evaluate sequence, after which the outcome may be used to update the model. Dawid~\cite{Dawid1984Prequential} introduced this principle, while Gama et al.~\cite{Gama2013StreamEvaluation} developed procedures for applying it to evolving data streams.

Incremental and prequential approaches have also been used in educational prediction~\cite{ZavaletaSanchez2022Dynamic,Casalino2023Incremental,Casalino2024Time,Casalino2025Evolving}. Zavaleta-S'anchez and Ben'itez-Guerrero~\cite{ZavaletaSanchez2022Dynamic} compared six incremental classifiers as learner performance records were processed sequentially.

Casalino et al.~\cite{Casalino2023Incremental} investigated conventional and fuzzy Hoeffding trees for pass/fail prediction using semester chunks. Later studies considered monthly, trimester and semester periods~\cite{Casalino2024Time} and used windowed prequential evaluation under different levels of outcome availability~\cite{Casalino2025Evolving}.

These applications focused mainly on final academic outcomes at the learner level or across monthly, trimester and semester periods.

\subsection{Temporal and Operational Validity}

Longitudinal prediction requires model development and evaluation to preserve the order in which information becomes available. Future information, temporal dependence and repeated observations can produce optimistic estimates when data are divided without accounting for time~\cite{Kaufman2012,Varma2006Bias,Roberts2017CrossValidation}. Forward-only feature construction and learner-aware partitioning can reduce information leakage~\cite{Paz2025CAPIRE}, while temporally ordered evaluation better represents prediction from past observations to future outcomes~\cite{Roberts2017CrossValidation}.

Conventional test--then--train evaluation generally assumes that an outcome is available before the next labelled case is processed~\cite{Gama2013StreamEvaluation}. When outcomes are delayed, prediction time must be distinguished from outcome availability, and model updating postponed until the outcome is known~\cite{Grzenda2020Delayed}.

Changes in participation, course structure, assessment schedules and learner composition may coincide with changes in outcome prevalence and predictive performance~\cite{Fei2015temporal,Liu2020temporal,Xia2023dropout,Labba2024towards}. Comparing performance, prevalence and prediction-score distributions across calendar periods can reveal instability hidden by aggregate measures.

Decision thresholds affect the numbers of flagged learners, missed disengagement cases and false alerts, as well as intervention workload. For imbalanced outcomes, precision--recall measures and the Matthews correlation coefficient complement ROC-based measures~\cite{He2009learning,Saito2015PR,Chicco2020MCC}. Reporting predictive performance and alert numbers therefore clarifies the implications of alternative threshold policies.

Previous studies have used recent activity, cumulative engagement and temporal participation patterns to represent different aspects of learner behaviour~\cite{Liang2016EdxDropout,Wen2020LocalCorrelation}. However, differences in how far ahead predictions are made and when models are updated may affect comparisons among feature sets.

The reviewed studies reveal a specific methodological gap. Educational stream studies have mainly applied incremental or prequential evaluation to final academic outcomes at the learner level or over monthly, trimester and semester periods. Next-week dropout studies have generally used retrospective partitions, independently trained week-specific models or models that remained fixed during evaluation. Existing studies therefore do not combine next-week inactivity-onset prediction with ordering learner--course observations by calendar time, waiting until outcomes become available, continuously updating the model and keeping it unchanged for predictions made at the same time. Within such a time-synchronous setting, questions also remain about the operational consequences of threshold selection, temporal stability and the comparative predictive value of different behavioural feature representations. These issues motivate evaluation under a common chronological framework.

\subsection{Research Questions}

These unresolved issues lead to the following research questions:

\begin{description}
	
	\item[\textbf{RQ1.}] to what extent can continuously updated models predict next-week student disengagement using only LMS activity available at the time of prediction under time-synchronous prequential evaluation?
	
	\item[\textbf{RQ2.}] how do alternative threshold policies affect threshold-dependent performance and alert volume?
	
	\item[\textbf{RQ3.}] how stable are predictive performance and predicted risk scores over calendar time, and what patterns of change emerge?
	
	\item \textbf{RQ4.} how does predictive performance vary across different representations of student behaviour during model selection?

\end{description}

\section{Methodology}
\label{sec:methodology}
\begin{figure*}[!t]
    \centering
    \resizebox{0.80\textwidth}{!}{%
        \rotatebox{90}{%
            \includegraphics[
                trim=1 1 1 1,
                clip
            ]{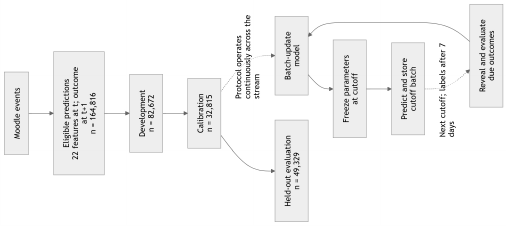}%
        }%
    }
    \vspace{-1mm}
    \caption{Time-synchronous prequential evaluation framework for next-week inactivity-onset prediction.}
    \label{fig:methodology_framework}
\end{figure*}

Next-week inactivity onset was treated as a one-week-ahead streaming prediction task. Learner--course observations were ordered by calendar prediction time, and models were updated only with outcomes already available. Model development, calibration, threshold selection and held-out prequential evaluation used separate chronological periods. Figure~\ref{fig:methodology_framework} summarises the framework.

\subsection{Data Construction and Prediction Task}
\label{sec:data_target}

Anonymised Moodle records from undergraduate courses at a South African public university were analysed retrospectively. Ethical and institutional approval was obtained for secondary analysis. The 6,912,737 events recorded between February 2013 and August 2019 were aggregated into 1,172,894 learner--course--week observations from 5,020 learners, 179 courses and 30,610 learner--course sequences. Course-relative time was limited to 52 weeks.

Each learner--course sequence was expanded from Week~1 to its observed course endpoint. After checking against the original event log, weeks without activity were assigned zero events, resulting in 1,188,883 weekly observations. The recorded course start time matched the first observed event and provided the calendar anchor. Week~$t$ began $7(t-1)$ days after this anchor; its prediction was issued at week end, and its outcome became observable seven days later.

Let $E_{ict}$ denote the number of Moodle events generated by learner $i$ in course $c$ during relative week $t$, and let $T_c$ denote the observed course endpoint. The eligible prediction set and next-week inactivity target were defined as

\begin{equation}
\begin{aligned}
\mathcal{R}
&=
\left\{(i,c,t):E_{ict}>0,\;t<T_c\right\},\\
y_{ic,t+1}
&=
\mathbf{1}\!\left(E_{ic,t+1}=0\right),
\qquad (i,c,t)\in\mathcal{R}.
\end{aligned}
\end{equation}

Thus, predictions were made only after active weeks and when the following course week was observable. Learners re-entered the risk set after reactivation, while terminal course weeks were excluded because their outcomes fell outside the observation period. The modelling cohort comprised 164,816 predictions from 4,791 learners and 23,593 learner--course sequences across 179 courses.

For feature vector $\mathbf{x}{ict}$, each model estimated the next-week inactivity probability $\hat{p}{ic,t+1}=\Pr(y_{ic,t+1}=1\mid\mathbf{x}{ict})$. An alert was generated when $\hat{p}{ic,t+1}\geq\tau$, where $\tau$ was the decision threshold.

\subsection{Behavioural Predictors and Preprocessing}
\label{sec:feature_sets}

Twenty-two predictors available up to and including the current week were organised into three behavioural feature groups and their combined representation (Table~\ref{tab:feature_groups}).

\begin{table*}[t]
	\centering
	\caption{Behavioural feature groups used during model development.}
	\label{tab:feature_groups}
	\small
	\setlength{\tabcolsep}{5pt}
	\renewcommand{\arraystretch}{1.10}
	
	\begin{tabularx}{\textwidth}{@{}
			>{\raggedright\arraybackslash}p{3.0cm}
			>{\centering\arraybackslash}p{0.6cm}
			>{\raggedright\arraybackslash}X
			@{}}
		\hline
		\textbf{Feature group} & \textbf{$p$} & \textbf{Measures} \\
		\hline
		
		Current activity & 7 &
		Log-transformed counts of total events, views, submissions,
		attempts, downloads, posts, and other actions in the current week. \\
		\hline
		
		Cumulative and temporal & 8 &
		Week-to-week change in log events; positive-change indicator;
		mean and standard deviation of log events to date; active-week rate;
		log mean event count to date; and the mean and standard deviation
		of log events over the most recent four weeks. \\
		\hline
		
		Risk history & 7 &
		Log active-streak length; log preceding-inactivity length;
		log maximum inactivity run to date; inactive-week rate;
		log inactivity-spell count; reactivation after at least two
		inactive weeks; and log reactivation count. \\
		\hline
		
		All features & 22 &
		All current-activity, cumulative-and-temporal, and
		risk-history predictors. \\
		\hline
	\end{tabularx}
    \vspace{-18pt}
\end{table*}

Where indicated, event-count, run-length, rolling and reactivation variables were transformed using $\log(1+x)$. Cumulative and history measures were calculated separately within each learner--course sequence. Learner and course identifiers, course endpoints, prediction times and outcome-derived fields were used only to construct the sequences, timing and labels; they were not predictors.

Predictors were standardised using the 41,342 warm-up rows. The transformation was then fixed, values were clipped to $[-8,8]$, and the same procedure was applied to all subsequent observations.

\subsection{Chronological and Prequential Design}
\label{sec:prequential_design}

Predictions were ordered by UTC prediction time and divided chronologically without separating records made at the same time: 82,672 development predictions (February 2013--February 2016), 32,815 calibration predictions (February 2016--April 2017) and 49,329 evaluation predictions (April 2017--May 2019).

Development comprised 41,342 warm-up observations, during which preprocessing was fitted and the streaming models began updating, and 41,330 tuning observations used for model selection. Candidate models were run from the start of development and selected using the 40,500 tuning outcomes available before calibration. Calibration and frozen-threshold selection used the 31,633 outcomes available by the end of the calibration period.

Whenever predictions were issued or outcomes became available, all newly available outcomes were incorporated in one batch before predictions were made at that time. The model then remained fixed for all such predictions. No outcome updated the model before becoming available. Evaluation outcomes could update later predictions but could not influence model development, calibration or frozen-threshold selection.

SGD-LR and OSELM used batch updates whose results did not depend on the order of records. ESN maintained a separate reservoir state for each learner--course sequence; the state advanced when a prediction was made, and the corresponding readout vector was retained until the outcome became available.

\subsection{Models and Development Selection}
\label{sec:streaming_models}

Three streaming models were evaluated. Stochastic gradient descent logistic regression (SGD-LR) estimated $\hat{p}=\sigma(w^{\top}x+b)$, where $\sigma$ was the sigmoid function, and updated $w$ and $b$ whenever a batch of outcomes became available. At update batch $k$, its learning rate was

\begin{equation}
\eta_k
=
\frac{\eta_0}{\sqrt{1+\delta k}},
\qquad \delta=10^{-3}.
\end{equation}

The online sequential extreme learning machine (OSELM) used a fixed random hidden layer,

\begin{equation}
h(x)
=
\tanh\left(\gamma(xW+b_h)\right),
\end{equation}

where $W$ and $b_h$ were the fixed hidden weights and biases and $\gamma$ was the input scale. Its regularised recursive least-squares (RLS) readout used outcomes encoded as $-1$ and $+1$; scores were mapped to $(0,1)$ using the sigmoid function~\cite{Huang2006ELM,Liang2006OSELM}.

The echo state network (ESN) maintained a reservoir state for each learner--course sequence:

\begin{equation}
s_{ic,t}
=
(1-\alpha)s_{ic,t-1}
+
\alpha\tanh\left(
W_{\mathrm{in}}x_{ic,t}
+
W_{\mathrm{res}}s_{ic,t-1}
+
b_r
\right),
\end{equation}

where $\alpha$ was the leak rate, $W_{\mathrm{in}}$ the input matrix, $W_{\mathrm{res}}$ the recurrent matrix and $b_r$ the reservoir bias. The RLS readout $z_{ic,t}=[1,x_{ic,t},s_{ic,t}]$ combined an intercept, current inputs and the updated state; it was updated after the outcome became available. The recurrent matrix was scaled so that its spectral radius, the largest absolute eigenvalue of $W_{\mathrm{res}}$, matched the selected value~\cite{Jaeger2001echo,Lukosevicius2009reservoir}.

For OSELM and ESN, the inverse information matrix was initialised as $P_0=(\lambda I)^{-1}$, where $\lambda$ was the selected ridge penalty and $I$ the identity matrix. Their hidden or reservoir weights remained fixed, and only the readout weights were updated~\cite{Huang2006ELM,Budiman2016adaptive}.

Three temporal reference scores were calculated: global and course-specific rates based on previously available outcomes, and an active-persistence rule that predicted continued activity. The two rates used Beta$(1,1)$ smoothing, equivalent to adding one prior count to each outcome class.

All candidates were compared on the same 40,500 tuning observations. Average precision (AP) was the primary selection measure, followed by ROC-AUC, with lower complexity used to break ties. Table~\ref{tab:selected_model_settings} gives the search spaces and selected configurations. OSELM and ESN used a random seed of 2026.

\begin{table}[t]
	\centering
	\caption{Development search spaces and selected model configurations.}
	\label{tab:selected_model_settings}
	\small
	\setlength{\tabcolsep}{4pt}
	\renewcommand{\arraystretch}{1.10}
	
	\begin{tabularx}{\textwidth}{
			>{\raggedright\arraybackslash}p{1.3cm}
			>{\raggedright\arraybackslash}X
			>{\raggedright\arraybackslash}X
		}
		\hline
		\textbf{Model} &
		\textbf{Development search} &
		\textbf{Selected configuration} \\
		\hline
		
		SGD-LR &
		Four feature sets; $\eta_0\in\{0.01,0.03,0.10\}$;
		$\lambda\in\{10^{-4},10^{-3},10^{-2}\}$. &
		All 22 features; $\eta_0=0.01$;
		$\lambda=0.01$; $\delta=0.001$. \\
		
		\hline
		OSELM &
		Four feature sets; hidden units $\in\{16,32,64\}$;
		ridge $\in\{0.1,1,10\}$. &
		All 22 features; 64 hidden units;
		ridge $=10$; input scale $=1$. \\
		
		\hline
		ESN &
		Four feature sets; reservoir units $\in\{32,64\}$;
		spectral radius $\in\{0.5,0.9\}$;
		leak rate $\in\{0.3,0.7\}$;
		ridge $\in\{1,10\}$. &
		All 22 features; 64 reservoir units;
		spectral radius $=0.5$; leak rate $=0.7$;
		ridge $=10$; input scale $=1$. \\
		\hline
	\end{tabularx}
    \vspace{-18pt}
\end{table}

\subsection{Calibration, Threshold Policies, and Performance Measures}
\label{sec:evaluation_measures}

After model selection, each model was replayed through the complete stream. A logistic calibrator was fitted to the raw scores from the 31,633 eligible calibration observations to produce calibrated probabilities:

\begin{equation}
\hat{p}^{\mathrm{cal}}
=
\sigma\left(
a+b\,\operatorname{logit}(\hat{s})
\right),
\end{equation}

where $\hat{s}$ was the model score before calibration, clipped to $[\varepsilon,1-\varepsilon]$ before applying $\operatorname{logit}(\hat{s})=\log[\hat{s}/(1-\hat{s})]$; $\varepsilon$ was the clipping constant, and $a$ and $b$ were estimated from the calibration data. The calibrator was then fixed for evaluation.

A frozen threshold was selected separately for each model by maximising the Matthews correlation coefficient (MCC) on the calibrated scores over the grid $\mathcal{T}=\{0.05,0.10,\ldots,0.95\}$:

\begin{equation}
\tau^{*}
=
\arg\max_{\tau\in\mathcal{T}}
\operatorname{MCC}\left(
y,
\mathbf{1}\left[\hat{p}^{\mathrm{cal}}\geq\tau\right]
\right).
\end{equation}

The resulting thresholds---0.45 for SGD-LR, 0.35 for OSELM and 0.40 for ESN---defined the primary model comparison. Four policies were applied to the same calibrated evaluation scores: fixed thresholds of 0.30 and 0.50, the frozen threshold and a rolling MCC threshold. The rolling threshold was updated only after the outcomes for a completed batch became available and was applied to the next batch. Model parameters and calibrated scores remained unchanged across policies.

Discrimination was assessed using average precision (AP) and the area under the receiver operating characteristic curve (ROC-AUC); AP was calculated using \texttt{average\_precision\_score}. Probability accuracy was assessed using the Brier score, the mean squared difference between predicted probabilities and observed outcomes. Threshold-dependent measures comprised MCC, F1-score, precision, recall, specificity, balanced accuracy, accuracy and confusion-matrix counts. The alert rate was the proportion of predictions classified as alerts. MCC was the primary decision measure because it incorporates all four confusion-matrix cells under class imbalance~\cite{Chicco2020MCC}.

\subsection{Calendar-Time Stability and Score Drift}
\label{sec:temporal_stability}

Calendar-time stability was evaluated using held-out calibrated probabilities and model-specific frozen thresholds. Prediction times were grouped into Monday--Sunday UTC weeks; weeks without observations were retained as gaps.

For each observed week and model, AP, ROC-AUC, Brier score, MCC, F1-score, precision, recall, specificity, balanced accuracy, prevalence, mean calibrated probability and alert rate were calculated. Weekly measures were summarised by their unweighted mean, standard deviation, median, interquartile range, minimum and maximum. Ordinary least-squares slopes and Spearman correlations with elapsed week estimated linear and monotonic trends, respectively, excluding missing values by measure. AP, ROC-AUC, MCC and balanced accuracy were omitted for single-class weeks. Precision, recall, specificity and F1-score were omitted when their denominators were zero; MCC was retained as zero when both outcome classes were present but predictions were constant.

Weekly calibrated-probability distributions were compared with their calibration distributions. Up to ten fixed quantile bins were derived from calibration probabilities, with duplicate boundaries removed; $B$ was the effective bin count. The population stability index (PSI) and Jensen--Shannon divergence measured changes in bin proportions, whereas the two-sample Kolmogorov--Smirnov statistic measured the maximum separation between empirical cumulative distributions. Bin counts were smoothed by $10^{-6}$ before normalisation. Drift statistics were reported only for weeks with at least $\max(30,5B)$ observations and interpreted as model-score drift indicators, not evidence of causal concept drift.

Experiments used Python 3.12.7, NumPy 1.26.4, pandas 2.2.3, SciPy 1.16.0 and scikit-learn 1.7.0. The preprocessing transformation, selected feature sets, model configurations, calibrators and frozen thresholds were fixed before evaluation. Integrity checks verified chronological ordering, eligibility, delayed outcome use, valid probabilities, deterministic model construction and locked settings.

\section{Results and Discussion}
\label{sec:results}

The held-out evaluation comprised 49,329 one-week-ahead predictions, including 18,588 inactivity-onset cases (37.68\%). Calibration used 31,633 observations with a prevalence of 33.38\%. Across models, sigmoid calibration reduced the Brier score from 0.194--0.200 to 0.182--0.188 and log loss from 0.573--0.590 to 0.546--0.560. The frozen MCC thresholds were 0.45 for SGD-LR, 0.35 for OSELM and 0.40 for ESN.

\begin{table}[t]
	\centering
	\caption{Held-out performance at model-specific frozen thresholds ($n=49{,}329$; prevalence $=37.68\%$); ESN had the highest point estimates for AP, ROC-AUC, MCC and balanced accuracy.}
	\label{tab:final_model_performance}
	\scriptsize
	\setlength{\tabcolsep}{1.5pt}
	\renewcommand{\arraystretch}{1.12}
	
	\begin{tabular*}{\textwidth}{
			@{\extracolsep{\fill}}
			lccccccccc
			@{}
		}
		\hline
		\textbf{Model} &
		\textbf{$\tau$} &
		\textbf{AP} &
		\textbf{\shortstack{ROC-\\AUC}} &
		\textbf{MCC} &
		\textbf{F1} &
		\textbf{Precision} &
		\textbf{Recall} &
		\textbf{Specificity} &
		\textbf{\shortstack{Balanced\\accuracy}} \\
		\hline
		SGD-LR & 0.45 & 0.665 & 0.755 & 0.396 &
		0.603 & 0.656 & 0.557 & 0.823 & 0.690 \\
		
		OSELM & 0.35 & 0.670 & 0.766 & 0.405 &
		0.643 & 0.601 & 0.692 & 0.722 & 0.707 \\
		
		ESN & 0.40 & 0.679 & 0.774 & 0.423 &
		0.636 & 0.648 & 0.625 & 0.795 & 0.710 \\
		\hline
	\end{tabular*}
    \vspace{-17pt}
\end{table}

Aggregate discrimination was similar across models: AP ranged from 0.665 to 0.679, exceeding the no-skill reference of 0.377, and ROC-AUC from 0.755 to 0.774 (Table~\ref{tab:final_model_performance}). Differences were small and not statistically tested.

The frozen thresholds revealed clearer operating differences. OSELM had the highest recall (0.692) and F1-score (0.643); SGD-LR had the highest precision (0.656) and specificity (0.823) but the lowest recall (0.557); and ESN had the highest MCC (0.423) and balanced accuracy (0.710). Thus, OSELM identified more cases, SGD-LR excluded more non-cases and ESN achieved the strongest balance.

\begin{table*}[!t]
	\centering
	\caption{Threshold-policy performance; lower thresholds increased recall and alert rate, while rolling thresholds did not improve aggregate MCC over frozen thresholds. For rolling policies, $\tau$ is the observation-weighted mean threshold.}
	\label{tab:threshold_policies}
	\footnotesize
	\setlength{\tabcolsep}{9pt}
	\renewcommand{\arraystretch}{1.08}
	\begin{tabular}{llcccc}
		\hline
		\textbf{Model} &
		\textbf{Policy} &
		\textbf{$\tau$} &
		\textbf{MCC} &
		\textbf{Recall} &
		\textbf{Alert rate (\%)} \\
		\hline
		SGD-LR & Fixed       & 0.300 & 0.346 & 0.736 & 51.37 \\
		SGD-LR & Fixed       & 0.500 & 0.386 & 0.489 & 26.90 \\
		SGD-LR & Frozen MCC  & 0.450 & 0.396 & 0.557 & 32.01 \\
		SGD-LR & Rolling MCC & 0.450 & 0.387 & 0.551 & 31.86 \\
		\hline
		OSELM & Fixed       & 0.300 & 0.392 & 0.738 & 48.64 \\
		OSELM & Fixed       & 0.500 & 0.400 & 0.545 & 30.75 \\
		OSELM & Frozen MCC  & 0.350 & 0.405 & 0.692 & 43.39 \\
		OSELM & Rolling MCC & 0.354 & 0.405 & 0.687 & 42.99 \\
		\hline
		ESN & Fixed       & 0.300 & 0.392 & 0.750 & 49.84 \\
		ESN & Fixed       & 0.500 & 0.417 & 0.515 & 27.52 \\
		ESN & Frozen MCC  & 0.400 & 0.423 & 0.625 & 36.34 \\
		ESN & Rolling MCC & 0.395 & 0.423 & 0.630 & 36.77 \\
		\hline
	\end{tabular}
    \vspace{-12pt}
\end{table*}

Threshold choice substantially affected recall and alert volume (Table~\ref{tab:threshold_policies}). At a threshold of 0.30, recall ranged from 0.736 to 0.750 and alert rates from 48.64\% to 51.37\%; at 0.50, recall fell to 0.489--0.545 and alert rates to 26.90\%--30.75\%.

At the frozen thresholds, alert rates were 32.01\% for SGD-LR, 43.39\% for OSELM and 36.34\% for ESN. Rolling adaptation reduced SGD-LR MCC from 0.396 to 0.387 and left OSELM and ESN unchanged to three decimal places; changes in recall and alert rate were also small. Thus, rolling adaptation did not improve aggregate MCC over the frozen thresholds.

The held-out predictions spanned 65 observed Monday--Sunday weeks within 108 calendar weeks. Fifty-nine weeks contained both outcome classes, permitting AP and MCC estimation, and 55 met the 50-observation minimum for drift statistics.

Mean weekly AP was 0.753 (SD $=0.126$) for SGD-LR, 0.744 (SD $=0.119$) for ESN and 0.734 (SD $=0.140$) for OSELM. Slopes ranged from $-0.0008$ to $-0.0006$ per calendar week, with no significant Spearman association between AP and time ($p\geq 0.457$). A common trough occurred during the week beginning 2 April 2018, when AP fell to 0.412, 0.391, and 0.359 across 1,478 predictions.

\begin{figure*}[!t]
	\centering
	\includegraphics[width=0.96\textwidth]{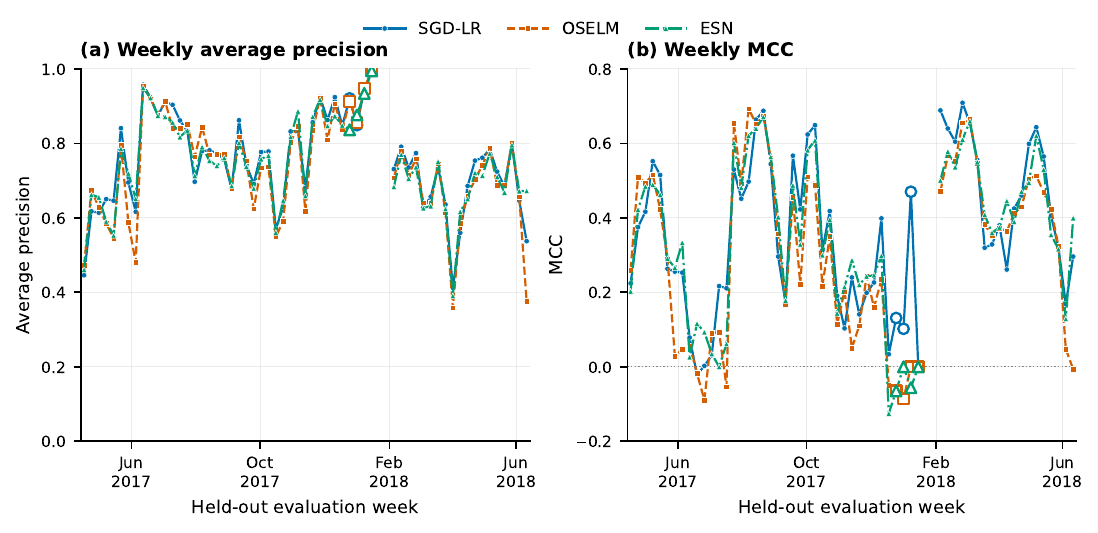}
	\caption{Weekly held-out AP (a) and MCC (b) for the three models; MCC varied more across weeks than AP, with no significant temporal trend in either measure. Breaks indicate non-estimable weeks; hollow markers denote $n<50$.}
	\label{fig:weekly_ap_mcc}
    \vspace{-18pt}
\end{figure*}

Mean weekly MCC was 0.364 (SD $=0.206$) for SGD-LR, 0.344 (SD $=0.212$) for ESN and 0.307 (SD $=0.234$) for OSELM (Figure~\ref{fig:weekly_ap_mcc}), with no significant Spearman association with time ($p\geq 0.140$). Mean weekly alert rates were 0.464, 0.515 and 0.584, respectively. These unweighted weekly estimates give equal weight to weeks, whereas aggregate estimates give equal weight to predictions.

\begin{figure*}[!t]
	\centering
	\includegraphics[width=0.96\textwidth]{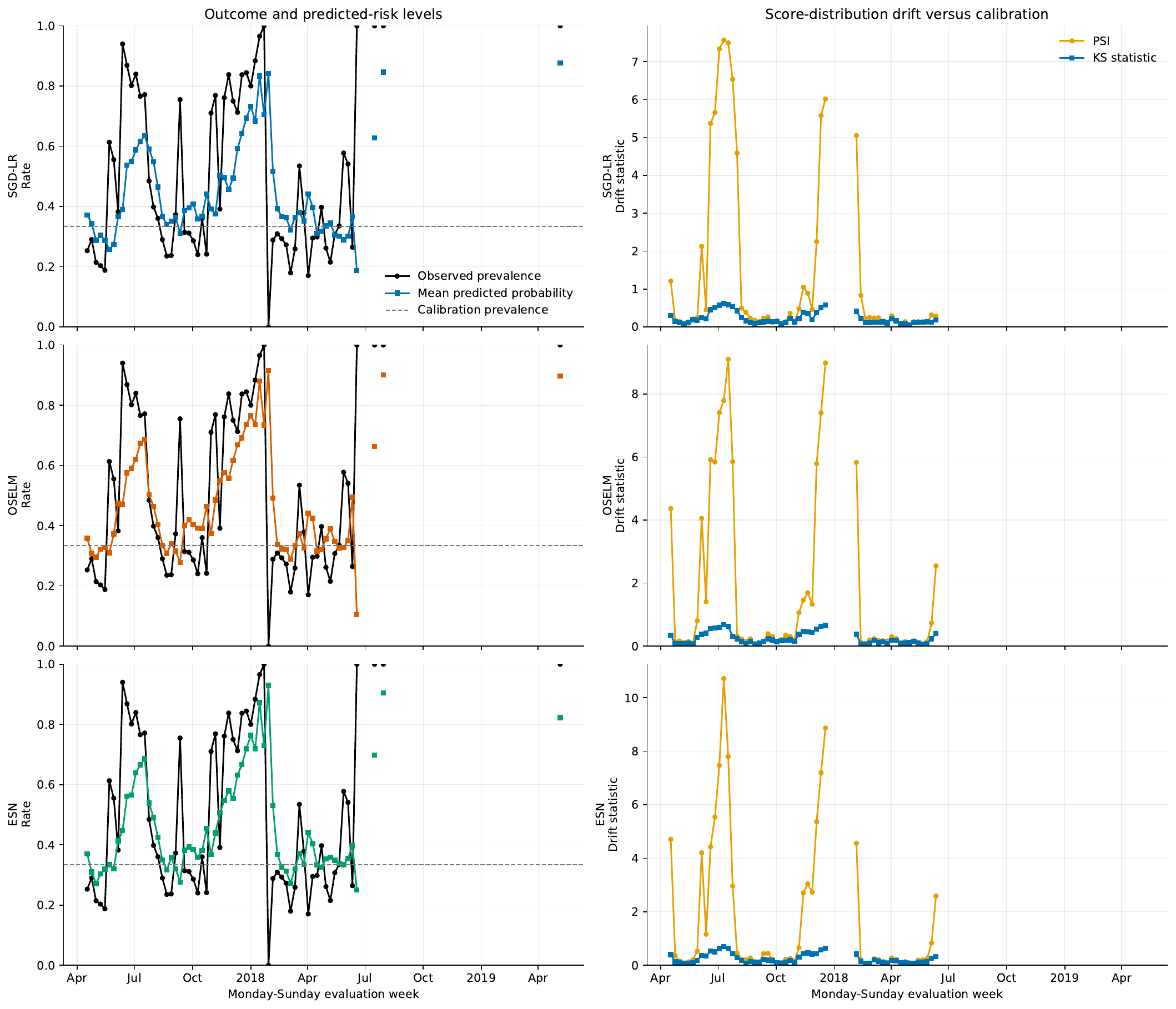}
	\caption{Weekly prevalence, mean calibrated probability and probability-distribution drift relative to calibration; drift was episodic and broadly aligned across models, with the largest shifts in July and December 2017. Breaks indicate weeks with $n<50$.}
	\label{fig:weekly_drift}
	\vspace{-18pt}
\end{figure*}

Inactivity-onset prevalence was 0.334 during calibration and 0.377 during evaluation, with no significant monotonic weekly trend. Across models, median PSI ranged from 0.234 to 0.247 and median KS from 0.145 to 0.185; maximum PSI ranged from 7.569 to 10.720 and maximum KS from 0.615 to 0.697 (Figure~\ref{fig:weekly_drift}). JJensen--Shannon divergence showed a similar episodic pattern.

Weekly AP, MCC and calibrated-probability distributions varied without sustained deterioration. Greater variation in MCC than AP suggests that threshold-dependent classification was less stable across weeks than ranking performance. PSI and KS indicate changes relative to the calibration distribution but do not establish concept drift.

\begin{table*}[!t]
	\centering
	\caption{Model-selection performance across four feature sets; the combined 22-feature set achieved the highest AP and ROC-AUC and lowest raw Brier score for all models.}
	\label{tab:development_feature_sets}
	\footnotesize
	\setlength{\tabcolsep}{7pt}
	\renewcommand{\arraystretch}{1.08}
	\begin{tabular}{llcccc}
		\hline
		\textbf{Model} &
		\textbf{Feature set} &
		\textbf{\shortstack{No.\\features}} &
		\textbf{AP} &
		\textbf{ROC-AUC} &
		\textbf{Raw Brier} \\
		\hline
		SGD-LR & Current                 & 7  & 0.593 & 0.711 & 0.239 \\
		SGD-LR & Cumulative and temporal & 8  & 0.696 & 0.759 & 0.199 \\
		SGD-LR & Risk history            & 7  & 0.664 & 0.727 & 0.217 \\
		SGD-LR & All features            & 22 & 0.704 & 0.769 & 0.192 \\
		\hline
		OSELM & Current                 & 7  & 0.596 & 0.725 & 0.213 \\
		OSELM & Cumulative and temporal & 8  & 0.684 & 0.759 & 0.205 \\
		OSELM & Risk history            & 7  & 0.663 & 0.739 & 0.210 \\
		OSELM & All features            & 22 & 0.700 & 0.774 & 0.201 \\
		\hline
		ESN & Current                 & 7  & 0.654 & 0.753 & 0.207 \\
		ESN & Cumulative and temporal & 8  & 0.684 & 0.764 & 0.204 \\
		ESN & Risk history            & 7  & 0.666 & 0.743 & 0.209 \\
		ESN & All features            & 22 & 0.705 & 0.780 & 0.199 \\
		\hline
	\end{tabular}
    \vspace{-8pt}
\end{table*}

Among the individual feature sets, cumulative-and-temporal features achieved the highest AP for each model: 0.696 for SGD-LR and 0.684 for both OSELM and ESN (Table~\ref{tab:development_feature_sets}). Current activity had the lowest AP throughout. Risk history improved AP over current activity but not consistently across the other metrics.

Relative to the cumulative-and-temporal set, the combined set increased AP by 0.008 for SGD-LR, 0.016 for OSELM and 0.020 for ESN. These gains supported its selection for the locked evaluation.

The time-synchronous evaluation exposed weekly variation hidden by aggregate estimates, supporting the need to preserve the order of predictions, outcome availability and model updating~\cite{Dawid1984Prequential,Gama2013StreamEvaluation}. Applied to next-week student disengagement, the framework extends sequential educational evaluation beyond final outcomes and coarser temporal windows~\cite{ZavaletaSanchez2022Dynamic,Casalino2023Incremental,Casalino2024Time,Casalino2025Evolving}.

Overall, discrimination differed little across models, and the combined feature set yielded modest gains during model selection. Threshold choice produced larger differences in recall and alert volume, whereas rolling adaptation did not improve aggregate MCC. Weekly performance and calibrated-probability distributions varied without sustained deterioration.

\section{Conclusion}
\label{sec:conclusion}

This study developed and evaluated a time-synchronous prequential framework for predicting next-week student disengagement, operationalised as the onset of inactivity, from Moodle LMS logs. The framework preserved the calendar order of prediction, outcome observation and model updating, reflecting the information available at each prediction point.

The models achieved similar discrimination but differed in their operating trade-offs. OSELM detected more inactivity-onset cases, SGD-LR generated more selective alerts, and ESN produced the strongest overall point estimates. Threshold choice substantially affected case detection and alert volume, whereas rolling adaptation did not improve performance relative to calibration-fixed thresholds. Cumulative-and-temporal behaviour was the strongest individual feature representation, with modest gains from combining all feature groups.

Weekly performance and score distributions varied without sustained deterioration, showing that aggregate metrics alone do not fully characterise streaming early-warning performance. Evaluation should therefore consider discrimination, calibration, threshold-dependent performance, alert burden and calendar-time variation. Model and threshold selection should reflect the intervention's purpose and capacity.

The study used historical Moodle records from one institutional setting. Future work could evaluate the framework in other learning environments and relate threshold selection to intervention capacity.

\begin{credits}

\subsubsection{\discintname}
The authors have no competing interests to declare that are relevant to the content of this article.
\end{credits}
%
%
%
%
%
%
\bibliographystyle{splncs04}
\bibliography{references}

\end{document}